\documentclass{article}

\usepackage{arxiv}

\usepackage[utf8]{inputenc} 
\usepackage[T1]{fontenc}    
\usepackage{hyperref}       
\usepackage{url}            
\usepackage{booktabs}       
\usepackage{amsfonts}       
\usepackage{nicefrac}       
\usepackage{microtype}      
\usepackage{lipsum}		
\usepackage{graphicx}
\usepackage{natbib}
\usepackage{doi}
\usepackage{mathtools} 
\usepackage{booktabs} 
\usepackage{tikz} 
\usepackage{amsmath}
\usepackage{amssymb}
\usepackage{makecell}
\usepackage[linesnumbered,ruled,vlined]{algorithm2e}
\usepackage{float}

\title{Nonlinear Hawkes Process with Gaussian Process Self Effects}

\author{Noa Malem-Shinitski \\
	Institute of Mathematics\\
    University of Potsdam\\
    Potsdam, Germany \\
	\texttt{malem@uni-potsdam.de} \\
	\And
	C\`{e}sar~Ojeda \\
	Artificial Intelligence Group\\
    Technische Universität Berlin\\
    Berlin, Germany \\
	\And
	Manfred Opper \\
	Artificial Intelligence Group\\
    Technische Universität Berlin\\
    Berlin, Germany \\
}

\date{}

\renewcommand{\shorttitle}{\textit{arXiv} Template}

\hypersetup{
pdftitle={A template for the arxiv style},
pdfsubject={q-bio.NC, q-bio.QM},
pdfauthor={David S.~Hippocampus, Elias D.~Striatum},
pdfkeywords={First keyword, Second keyword, More},
}

\begin{document}
\maketitle

\begin{abstract}
	Traditionally, Hawkes processes are used to model time--continuous point processes with history dependence. Here we propose an extended model where the self--effects are of both excitatory and inhibitory type and follow a Gaussian Process. Whereas previous work either relies on a less flexible parameterization of the model, or requires a large amount of data, our formulation allows for both a flexible model and learning when data are scarce. We continue the line of work of Bayesian inference for Hawkes processes, and our approach dispenses with the necessity of estimating a branching structure for the posterior, as we perform inference on an aggregated sum of Gaussian Processes. Efficient approximate Bayesian inference is achieved via data augmentation, and we describe a mean--field variational inference approach to learn the model parameters. To demonstrate the flexibility of the model we apply our methodology on data from three different domains and compare it to previously reported results.
\end{abstract}

\section{Introduction}

Sequences of self exciting, or inhibiting, temporal events are frequent footmarks of natural phenomena: Earthquakes are known to be temporally clustered as aftershocks are commonly triggered following the occurrence of a main event \citep{ogata}; in social networks, the propagation of news can be modeled in terms of information cascades over the edges of a graph \citep{SEISMIC}; and in neuronal activity, the occurrence of one spike may increase or decrease the probability of the occurrence of the next spike over some time period \citep{dayan2001theoretical}.

Traditionally, sequences of events in continuous time are modeled by Point processes, of which Cox processes \citep{cox1955some}, or doubly stochastic processes, use a stochastic process for the intensity function, which depends only on time and is not effected by the occurrences of the events. The Hawkes process \citep{HAWKES} extends the Cox process to capture phenomena in which the past events affects future arrivals, by introducing a memory dependence via a memory kernel. When incorporating dependence of the process on its own history, due to the superposition theorem of point process, new events will depend on either an exogenous rate, which is independent of the history, or an endogenous rate from past arrivals. This results in a branching structure, where new events that originate from previous events can be seen as "children" of the past events.


Originally, the dependence on the history in the Hawkes process is assumed to be self excitatory, and the memory kernel is parameterized by an exponential or power law decay, which results in a model with low flexibility. Furthermore, assuming only excitatory relation between the events, does not hold for other phenomena we wish to model. For example, inhibitory effects between neurons \citep{maffei2004selective}, and even self--inhibition \citep{smith2002self}, are crucial for regulating the neuronal activity. Thus, the memory kernel should also include inhibitory relations between the events and by doing so the intensity may become negative. To ensure that the intensity function is non--negative, a nonlinear link function is applied on the memory kernel, and the resulting model is often referred to as a Nonlinear Hawkes process \citep{bremaud1996stability, zhu2013central, truccolo2016point}.

In this work we present a \textit{Nonlinear Hawkes process with Gaussian Process Self--effects} (NH-GPS) which extends the class of Nonlinear Hawkes processes. We choose a non--parametric approach which avoids the limiting parameterization of the memory kernel and the background rate. We assume a Gaussian Process (GP) prior on the exogenous events intensity and on the memory kernel, which allows also for an inhibitory effect between the events. To ensure that the intensity function is non--negative we use the Sigmoid link function. This modeling approach is not only descriptive, but also allows us to obtain a fast inference procedure. The history of self--effects defines an aggregated Gaussian process, and we perform the inference directly on this aggregation rather than obtaining a posterior over each self effect. 


While highly flexible approaches to modeling the intensity function of nonlinear Hawkes processes have been presented before, they mainly rely deep neural network solutions \citep{jia2019neural, xiao2017wasserstein}.  These approaches are hindered by the necessity of large datasets. However, our methodology retains the modeling flexibility due to the non--parametric nature of Gaussian processes, while being able to perform well when data are scarce.

Our main contributions are the following
\begin{itemize}
  \item We present a new flexible non--parametric approach to the nonlinear Hawkes process, which allows us to perform inference when data are scarce.
  \item We derive an efficient variational inference scheme to the model.
\end{itemize}

\paragraph{Outline} In Section~\ref{sec:related} we discuss related work and describe how it relates to ours. In Section~\ref{sec:model} we describe the NH--GPS model and emphasize how its structure allows for efficient Bayesian inference. In Section~\ref{sec:inference} we describe the augmentation scheme and derive the mean--field variational inference algorithm. In Section~\ref{sec:exp} we present the results of our experiments both on synthetic data and different real world examples, and compare to existing work when possible. In Section~\ref{sec:con} we conclude by discussing our work and future research directions.

\section{Related Work} \label{sec:related}

Bayesian approaches to Cox Processes model the intensity with a Gaussian process prior, which is then passed through a link function to ensure its positivity. A common choice of the link function is the exponential or the quadratic functions \citep{hensman2015mcmc, lloyd2015variational}. Another choice, which is more relevant to our wok, is the sigmoid link function, resulting in the \emph{sigmoidal Gaussian Cox process}. Inference in this model was first done via Markov Chain Monte Carlo  \citep{adams2009tractable} and later via variational inference \citep{donner2018efficient}.

As for the Hawkes process, first attempts to perform Bayesian inference relied on the definition in terms of a marked Poisson cluster process and identifying the branching structure of the self--excitation \citep{HawkesBayes}. Alternative approaches to inference in temporal point processes and Hawkes processes include directly maximizing the Likelihood function via stochastic gradient descent \citep{du2016recurrent} or re--modeling the intensity function as a Recurrent Neural Network (RNN) \citep{mei2017neural, jing2017neural, zuo2020transformer}. The latter often require training over very large datasets to produce reliable results.

Other than RNN, another highly flexible approach to estimating the intensity function of the Hawkes process relies on GP priors \citep{zhang2018efficient, zhou2019efficient, zhang2020variational}. Recent adaptation of this approach is the model described in \citet{zhou2020efficient}. Similarly to the model described in our work, the authors avoid the limiting parameterization of the memory kernel by using GPs. Differently to our work, \citet{zhou2020efficient} remain in the linear Hawkes process regime and assume that the effects of past events are assumed to be only excitatory, whereas our approach allows both excitatory and inhibitory effects.

Noticeable variations of the nonlinear Hawkes process are the Isotonic Hawkes Process \citep{wang2016isotonic} and the Mutually Regressive Point Process \citep{apostolopoulou2019mutually}. In the first variation, although the model allows for both inhibitory and excitatory self--effects, there is a relation of either--or between them. Thus, the entire history may be either inhibiting or exciting. In contrast, the GP prior in our approach allows for the self-effects to change over time. Hence, one particular event may have inhibiting effect on the next one, but have an exciting effect on the one after that.

The second variation is more closely related to our work. The Mutually Regressive Point Process is designed to model neuronal spike trains. In this work, the classical self--excitatory Hawkes Process intensity function is augmented by a probability term. This term induces inhibition when it is close to zero. In a sense, this model includes two memory kernels -- one excitatory only which appears in the intensity function and another which can also induce inhibition in the augmenting probability term. In the current work, we achieve such flexibility of the self--effects in a simpler fashion by assuming the GP prior on the self effects. As mentioned before this also allows for the type of effect to change over time, which dows not appear in the work of \citet{apostolopoulou2019mutually}.

\section{Proposed Model} \label{sec:model}
\subsection{Classical Hawkes Process}
 Let $\mathcal{T}_T = [0,t] \in \mathbb{R}$. We define the counting measure $N(\mathcal{T}_t)$ as the number of arrivals in the sequence $\mathcal{H}_t =  \{T_1,...,T_{N(\mathcal{T}_t) } : T_i \in \mathcal{T}_t \wedge  T_{i-1} < T_{i} \}$ where $\mathcal{H}_t$ defines the history of the process until time $t$, and $T_i$ corresponds to the time of arrival $i$. For a temporal point process, the counting measure $N(\cdot)$ has an associated intensity defined as
\begin{align*}
\Lambda(t) = \lim_{\Delta t \rightarrow 0} \frac{\mathbb{E}[N(\mathcal{T}_{t + \Delta t}) - N(\mathcal{T}_t) | \mathcal{H}_t ]}{\Delta t}\mbox{.}
\end{align*}

The intensity function may depend on the history of the process. An example of such a process is the Hawkes process, or self exciting point process, \citep{Kingman} which  defines self excitations \citep{daley2007introduction} around \textit{exogenous events}.

Following \citet{HAWKES}, the intensity of the Hawkes process is defined by
\begin{align}
    \Lambda(t \vert \mathcal{H}_t) = s\left(t\right) + \sum_{t_n < t} g\left(t - t_n \right)\mbox{,}
\label{eq:hawkes_intensity}
\end{align}
where $s(t)$ is the  base intensity of exogenous arrivals and $g\left(t - t_n \right)$ is the memory kernel defining the change in the excitation value for each arrival. In the classical Hawkes process, only excitations are allowed and the memory kernel is usually of the form $g(t-t_n) = \beta e^{-\alpha(t-t_n)}$ for an exponentially decaying memory.


\subsection{Nonlinear Hawkes process with Gaussian Process Self--Effects} \label{sec:full_model}
In the classical Hawkes process, the memory kernel $g$ in Equation~\ref{eq:hawkes_intensity} must be non--negative, to prevent the intensity function from being negative. As a result the history of the model has only excitatory effect on future events. We are interested in a model that includes inhibition between events, and we release the constraint over $g$ so it can be negative, and define the following nonlinear intensity function
\begin{align} \label{eq:nonlinear}
    \Lambda&\left(t \right) = \lambda \sigma \left(\phi(t) \right) \\
    \sigma \left(\phi(t) \right) &= \frac{1}{1 + \exp{\left(-\phi \left(t\right)\right)}} \\
    \phi(t) = s\left(t\right) + &\sum_{t_n < t} g\left(t - t_n \right)\exp\left(-\alpha \left( t - t_n\right) \right). \label{eq:phi}
\end{align}
Here, we choose the sigmoid function to ensure that the intensity function $\Lambda\left( \cdot \right)$ is non--negative. $\lambda$ is the intensity bound and we refer to $\phi\left( \cdot \right)$ as the linear intensity function.

We explicitly add the exponential decay to enforce the forgetting constraint which is essential for most realistic processes. Although we choose here a specific parameterization of the memory decay, one can choose other forms of memory decay with minimal adaptation to the learning procedure of the model parameters.

To maximize the flexibility of the model we avoid any specific parameterization of the background rate or the memory kernel. Thus, rather than specifying a functional form for $s\left(t\right)$ and $g\left(t\right)$, we assume the following GP priors
\begin{align}
s &\sim GP\left(0, K^s \right) \\ \label{eq:background_gp}
    g &\sim GP\left(0, K^g \right) \\ \label{eq:memory_gp}
    K&\left(t_1, t_2 \right) = a \cdot \exp{\left(- \frac{\| t_1 - t_2\|^2}{\sigma^2}\right)}.
\end{align}
In this work we use the Radial Basis Function (RBF) kernel for the GP priors. This choice is not a constraint of the model -- one can choose any other kernel, and it will not effect the augmentation and inference processes described bellow.

Finally, we assume a prior distribution also on the upper intensity bound
\begin{align*}
    \lambda \sim Gamma\left(\alpha_0, \beta_0\right).
\end{align*}
and we identify the hyperparameters of the model as $\{\sigma_g, a_g, \alpha, \sigma_s, a_s\}$.

In this work we propose Bayesian inference for fitting the model to data. Due to the non--linearity over $\phi\left(\cdot \right)$ we are no longer able to easily utilize the branching structure of the Hawkes process which allowed for the estimation of $s\left( \cdot \right)$ and $g\left( \cdot \right)$ \citep{HawkesBayes, zhou2020efficient}. Thus, a natural solution is to perform the inference directly on $\phi\left(\cdot \right)$. 

Next, we identify the prior over the entire linear intensity $p\left(\phi\right)$. From Equation~\ref{eq:phi} we see that the linear intensity function $\phi$ is nothing but the sum of GPs, and as such it is also a GP
\begin{align*}
    &\phi \sim GP\left(0, \Tilde{K} \right) \\ 
    &\Tilde{K}_{lk} = K^s_ {lk} +  \sum_{t_i < t_l}\sum_{t_j < t_k} K^g_{t_l - t_i, t_k - t_j} \exp\left(-\alpha \left( t_l - t_i + t_k - t_j \right) \right).
\end{align*}


\subsubsection{Multivariate Model}
We propose an extension of the model to multiple dimensions. This is useful in applications where different types of events are observed, or the events originate from different processes that effect each other. We define an $R$--dimensional point process with intensity in dimension $r$
\begin{align*}
    &\Lambda^r\left(t\right) = \lambda^r\sigma\left(\phi^r\left(t \right) \right) \\
    &\phi^r\left(t \right) = s^r\left(t \right) + \sum_{m = 1}^R \sum_{t^m_n < t}g_{r,m}\left(t - t^m_n\right) \exp\left(-\alpha_{r,m} \left(t - t^m_n \right) \right)
\end{align*}
where $t_n^m$ is the time of event number $n$ of type $m$. We assume that every dimension has its own intensity bound $\lambda^k$ and background rate $s^k\left( \cdot\right)$. The different dimension interact with each other via the self effects term. $g_{k,l}\left( \cdot\right)$ defined the effects of events of type $l$ on events of type $k$. As in the univariate case, this effect may change over time.

Given the observations, the different dimensions are independent of each other and we can learn their parameters separately. Thus, in the following section we present the inference for the univariate model, and the extension to the multivariate case is straight forward.

\section{Inference} \label{sec:inference}
Conditioned on the intensity function $\Lambda\left(\cdot \right)$, the likelihood of observations $\{t_1, ... t_n \}$ from a Hawkes process is \citep{daley2003introduction}
\begin{align*}
\ell\left(\{t_1, ... t_n \}|\Lambda\left( \cdot \right)\right) = \exp\left\{ -\int^{T}_{0}\Lambda\left(t'\right)dt' \right\}\prod^N_{i=1}\Lambda(t_i).
\label{eq:like}
\end{align*}
Looking at the likelihood defined above, Equations~\ref{eq:phi} and \ref{eq:like}, implementing Bayesian inference for the model is not straightforward, due to the non-conjugate structure of the likelihood and prior. Similarly to previous work on Cox and Hawkes processes \citep{donner2018efficient, apostolopoulou2019mutually, zhou2020efficient}, we augment the model with auxiliary variables, which leads to a conditionally conjugated model with closed form solutions for Gibbs sampler and variational inference.

\subsection{Model Augmentation}
The first step we take in treating the likelihood function is using the P\'{o}lya-Gamma (PG) augmentation scheme. Following Theorem $1$ in \citet{polson2013bayesian}, we can rewrite the nonlinear intensity function as
\begin{align}
    &\sigma\left(\phi\left(t\right)\right) = \int_0^{\infty} e^{f\left(w, \phi\left(t\right) \right)}PG\left(w; 1,0\right)dw \\
    &f\left(w, \phi\left(t\right)\right) = -\frac{\phi\left(t\right)^2w}{2} + \frac{\phi\left(t\right)}{2} - \ln{2}. \label{eq:f}
\end{align}
As we augment each observation with a variable $w_n$ from a PG distribution, the joint likelihood of the observed events $\{t_n\}$ and PG variables $\{w_n\}$ is
\begin{align}\label{eq:with_wn}
   & p\left(\{t_n\}_{n=1}^N, \{w_n\}_{n=1}^N | \phi, \lambda \right) = \\ \nonumber
   &\exp\left(-\int^T_0\lambda \sigma \left(\phi \left( t\right)\right)dt\right) \cdot \prod^{N}_{n=1}\lambda e^{f\left(w_n, t_n\right)}PG\left(w_n; 1,0\right)
\end{align}
with
\begin{align}\label{eq:int_pg}
    &\exp\left\{-\int^T_0\lambda \sigma \left(\phi\left( t\right)\right) dt\right\} = \\ \nonumber
    &\exp\left(-\int_0^T \int_0^\infty \lambda PG\left( w;1, 0\right) \left( 1 - e^{f\left(w, -\phi\left(t\right) \right)}\right)dwdt\right).
\end{align}

Where we used $\sigma(t) = 1 - \sigma(-t)$. 

Next, we utilize the Campbell's theorem \citep{Kingman} which states that for a Poisson process $\Pi$ with intensity $\varphi$
\begin{align*}
    \mathbb{E}_{\varphi}&\left(\prod_{x \in \Pi} \exp\left(h\left(x \right) \right) \right) = \\ &\exp \left(- \int \left(1 - \exp\left(h\left(x\right) \right) \right)\varphi\left(x \right)dx \right).
\end{align*}
Looking at Equation~\ref{eq:int_pg} we identify $x = \left(t, w \right)$ and $\varphi\left(t, w \right) = \lambda PG \left(w \vert 1,0 \right)$ is the intensity of a marked Poisson process in $\mathcal{T}$ with marks $w \sim PG \left(0, 1\right)$. Further, we determine $h\left( x\right) = f\left(w, -\phi\left(t \right) \right)$. We can now rewrite the exponential in Equation~\ref{eq:with_wn} as 
\begin{align} \label{eq:with_augs}
    \exp\left\{-\int^T_0\lambda \sigma \left(\phi\left( t\right)\right) dt\right\} = \mathbb{E}_{\varphi} \left( \prod_{m=1}^Me^{f\left(\hat{w}_m, \hat{t}_m  \right)}\right)
\end{align}
for realizations $\{\hat{t}_m, \hat{w}_m \}_{m = 1}^M$.

We substitute Equation~\ref{eq:with_augs} into Equation~\ref{eq:with_wn} which results in the full augmented likelihood. Given the prior distribution over $\phi$ and $\lambda$, we can now write the model's posterior distribution as
\begin{align} \label{eq:aug_post}
    &p\left(\{\hat{t}_m, \hat{w}_m\}, \{w_n\}, \phi, \lambda|\{t_n \} \right) \propto \exp\left(-\lambda T\right) \times \\ \nonumber
    &\prod_{m=1}^M \lambda e^{f\left(\hat{w}_m, -\phi\left(\hat{t}_m\right) \right)} PG\left(w_m;1,0 \right) \times \\ \nonumber
    &\prod_{n=1}^N \lambda e^{f\left(w_n, \phi\left(t_n\right) \right)} PG\left(w_n;1,0 \right) \times p\left(\phi \right) p\left(\lambda \right).
\end{align}
To summarize, we augment the model with two sets of variables -- the PG variables $\{w_n\}$ which augment the actual realizations and the tuples $\{\hat{t}_m, \hat{w}_m\}$ which are the realizations and marks of the auxiliary marked Poisson process.

As mentioned above, we intend to learn directly the linear intensity function $\phi\left(\cdot\right)$. This allows us to utilize the efficient mean--field variational inference previously introduced in \citet{donner2018efficient} and \citet{donnerefficient}. Next, we go through the steps of the algorithm, and we refer the reader to the two papers mentioned above for further details. As a baseline we compare the performance of the variational inference algorithm to a Gibbs sampler. The details of the Gibbs sampler can be found in the Supplementary Material. 

\subsection{Variational Inference}
In variational inference \citep{jordan1999introduction, bishop2006pattern} we define a tractable distribution family and adapt it to approximate the posterior by maximizing the lower bound $\mathcal{L}(Q)$ defined below. This procedure minimizes the Kullback--Leibler divergence between the unknown posterior and the proposed approximating distribution. The posterior density is approximated by
\begin{align*}
&p\left(\{\hat{t}_m, \hat{w}_m\}, \{w_n\}, \phi, \lambda|\{t_n \} \right) \\ \nonumber
&\approx q_1\left(\phi,\lambda\right) q_2\left(\{w_n\}_{n=1}^{N}, \{\hat{t}_m, \hat{w}_m\}_{m=1}^{M}\right).
\end{align*}
This leads to the following lower bound on the evidence
\begin{align*}
\mathcal{L}(Q) = \mathbb{E}_{Q}\left[\log\left\{\frac{p\left(\{\hat{t}_m, \hat{w}_m\}, \{w_n\}, \phi, \lambda|\{t_n \} \right)}{q_1\left(\phi,\lambda\right) q_2\left(\{w_n\}_{n=1}^{N}, \{\hat{t}_m, \hat{w}_m\}_{m=1}^{M}\right)}\right\}\right]\mbox{.}
\end{align*}
Here $Q$ refers to the probability measure of the variational posterior. We can maximize the bound by alternating the maximization over each of the factors \citep{bishop2006pattern}. The optimal solution for each factor is
\begin{align}\label{eq:optimal}
&\log q^*_{1}\left(\phi,\lambda\right) = \\ \nonumber
&\mathbb{E}_{q_{2}\left(\{w_n\}_{n=1}^{N}, \{\hat{t}_m, \hat{w}_m\}_{m=1}^{M}\right)}[\log P(\{\hat{t}_m, \hat{w}_m\}, \{w_n\}, \phi, \lambda, \{t_n \})] \\
&\log q^*_{2}\left(\{w_n\}_{n=1}^{N}, \{\hat{t}_m, \hat{w}_m\}_{m=1}^{M}\right) = \\ \nonumber
&\mathbb{E}_{q_{1}\left(\phi,\lambda \right)}[\log P(\{\hat{t}_m, \hat{w}_m\}, \{w_n\}, \phi, \lambda, \{t_n \})].
\end{align}
Thus, to obtain the optimal distribution of one of the factors, one must calculate expectations of the logarithm of the joint distribution over the remaining factors in the approximation, resulting in an iterative algorithm. 

In the following subsections, we explicitly express the functional form of the optimal distributions, and  obtain the corresponding expectations required for updating the factors. The hyperparameters ($\{\sigma_g, a_g, \alpha, \sigma_s, a_s\}$) are learned via gradients update of the lower bound, we present details in the Supplementary Material.

\subsection{Optimal $q_1$}
We find that the optimal $q_1$ is factorized as
\begin{align*}
    q_1\left( \phi, \lambda\right) = q_1\left(\lambda\right) q_1\left(\phi\right)
\end{align*}

The first factor is identified as a Gamma distribution
\begin{align}\label{eq:vi_lambda}
    &q_1 \left(\lambda \right) = Gamma\left(\alpha, \beta \right) \\ \nonumber
    &\alpha = \alpha_0 + N + \int_{\mathcal{T} x \mathcal{W}} \Lambda_{q_2}\left(t, w\right)dtdw \\ \nonumber
    &\beta = \beta_0 + T
\end{align}
with known expectations.

\begin{figure*}
    \centering
    \includegraphics[width=0.99\linewidth]{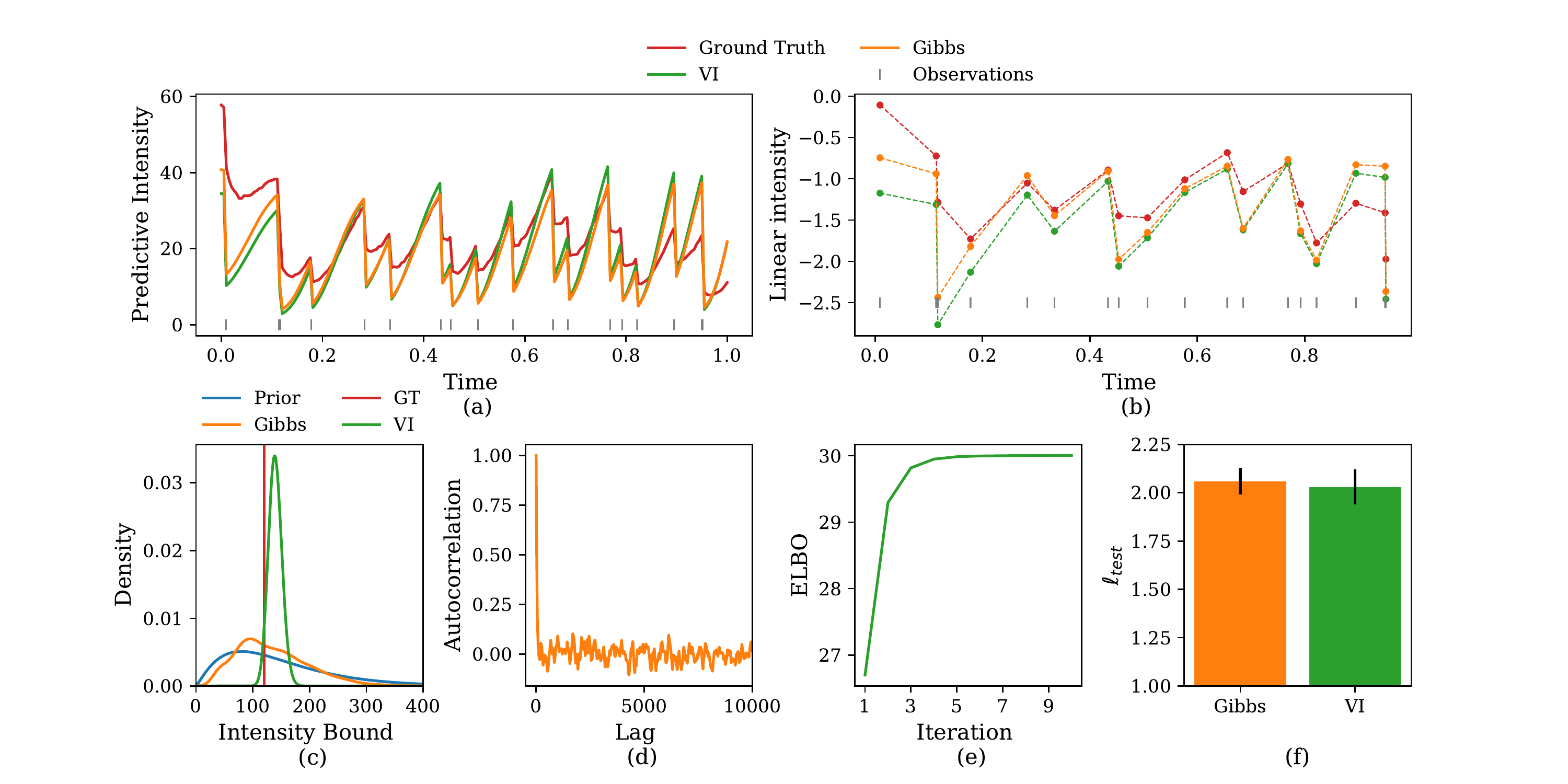}
  \caption{(a) Comparison of the ground truth predictive intensity and the one sampled from the VI and Gibbs inference. (b) Comparison of the ground truth linear intensity $\phi\left( \cdot\right)$ and the ones learned by the VI and Gibbs sampler. (c) Comparison of the Ground truth intensity bound and the one learned by the inference, and the prior distribution. (d) The autocorrelation of the intensity bound Gibbs samples. (e) The variational lower bound as a function of the algorithm iteration. (f) Comparison of the test log--likelihood of the Gibbs sampler and the VI.}\label{fig:gen_dat}
\end{figure*}

The optimal distribution for the second factor is of the form
\begin{align*}
    q_1^\star &\propto e^{- U\left( \phi \right) + \log p\left(\phi\right)} \\
    U(\phi) &= \frac{1}{2}\int A(t) \phi^2(t) dt - \int b(t)\phi(t)dt \\
    A(t) &= \sum_n\langle \omega_n \rangle_{q_2^\star}\delta(t - t_n) + \langle \omega(t) \rangle_{q_2^\star}\Lambda_{q_2^\star}(t) \\
    b(t) &= \sum_n\frac{1}{2}\delta(t - t_n) - \frac{1}{2}\Lambda_{q_2^\star}\left(t\right).
\end{align*}
Generally, the integrals above cannot be evaluated analytically. Thus, we resort to another variational approximation, where we approximate the likelihood term, by a distribution that depends only on a finite set of inducing point $\{c\}$, $\tilde{q}\left(\phi_c, \phi \right) = p\left(\phi \vert \phi_c \right) q\left(\phi_c\right)$  and the ELBO is
\begin{align*}
    \left\langle \log \frac{e^{-\log\langle U(\phi)\rangle_{p(\phi\vert \phi_c)}}p(\phi_c)}{\tilde{q}(\phi_c)}\right\rangle_{\tilde{q}}
\end{align*}
and we use the notation $\langle p\rangle_q = \mathbb{E}_q\left(p\right)$. The optimal $\tilde{q}\left(\phi_c\right)$ is given by
\begin{align*}
    \tilde{q}^\star(\boldsymbol{\phi}_c) \propto e^{-\log\langle U(\boldsymbol{\phi})\rangle_{p(\boldsymbol{\phi}\vert \boldsymbol{\phi}_c)}}p(\boldsymbol{\phi}_c).
\end{align*}
From here, using known results of conditional GPs and sparse variational GPs \citep{csato2001tap, titsias2009variational} we have
\begin{align} \label{eq:vi_gp_ind}
\tilde{q}^\star(\boldsymbol{\phi}_c) &= \mathcal{N}(\boldsymbol{\phi}_c\vert \boldsymbol{\mu}_c, \Sigma_c) \\ \nonumber
\Sigma_c &= \left[\int \boldsymbol{\kappa}(t)^\top A(t)\boldsymbol{\kappa}(t) dt + K_c^{-1}\right]^{-1}\\ \nonumber
\boldsymbol{\mu}_c &= \Sigma_c\left(\int b(t)\boldsymbol{\kappa}(t) dt\right)
\end{align}
with $K_c$ the covariance kernel between the inducing points, $\boldsymbol{\kappa}(t)=\boldsymbol{k}_c(t)^\top K_c^{-1}$ and $\boldsymbol{k}_c(t)$ is the kernel between the inducing points and another set of points (either the real data or the integration points). The mean and the variance of the sparse approximated GP are
\begin{align} \label{eq:vi_gp_dat}
\langle g(t)\rangle = &\boldsymbol{\kappa}(t)\boldsymbol{\mu}_c \\ \nonumber
\sigma^2(t) = & K(t,t) - \boldsymbol{\kappa}(t)^\top\boldsymbol{k}_c(t) + \boldsymbol{\kappa}(t)^\top \Sigma_c \boldsymbol{\kappa}(t)
\end{align}

\subsection{Optimal $q_2$}
Similarly to the previous section, we find that the optimal $q_2$ is factorized as
\begin{align*}
    q_2\left(\{w_n\}_{n=1}^{N}, \Pi\right)=q_2\left(\{w_n\}_{n=1}^{N}\right)q_2\left(\{\hat{t}_m, \hat{w}_m\}\right)
\end{align*}
Given Equation~\ref{eq:aug_post}, we define the first factor as
\begin{align*}
    q_2^\star(w_n) \propto \exp \left(-\frac{\langle \phi_n^2 \rangle_{q_1^\star}}{2}w_n\right) PG(w_n\vert 1, 0),
\end{align*}
which corresponds to a tilted PG distribution
\begin{align} \label{eq:vi_pg}
    q_2^\star(w_n) = PG\left(w_n\vert 1, \sqrt{\langle \phi_n^2\rangle_{q_1^\star}}\right).
\end{align}
with known expectations \citep{polson2013bayesian}.

The second factor takes the form
\begin{align*}
    & q_2^\star(\{\hat{t}_m, \hat{w}_m\}_{m=1}^{M}) \\ \nonumber
    &\propto \prod_{m=1}^M \exp \left(-\frac{\langle \phi_m\rangle_{q_1^\star}}{2} - \frac{\langle \phi_m^2\rangle_{q_1^\star}}{2}w_m\right)
    \cdot \exp \left(\left\langle\ln \lambda^\star\right\rangle_{q_1^\star} \right).
\end{align*}

It can be shown that this distribution corresponds to a Poisson process with intensity function
\begin{align}\label{eq:vi_lambda2}
    &\Lambda_{q_2}\left(\hat{t}, \hat{w}\right) \\ \nonumber
     &= \exp \left(\langle \ln \lambda \rangle_{q_1^\star} \right) \frac{\exp \left(-\frac{\langle \phi\rangle _{q_1^\star}}{2} \right)}{2 \cosh\left(\langle \phi^2\rangle _{q_1^\star} \right)} PG\left(w_m\vert 1, \sqrt{\langle \phi^2\rangle _{q_1^\star}} \right)
\end{align}
where to simplify the notation we write $\phi$ instead of $\phi \left(\hat{t}\right)$.

We summarize the algorithm and discuss the hyper parameters learning in the Supplementary Material.

\section{Experiments}\label{sec:exp}
All the algorithm and experiments for this work are implemented in Python and are available online. For further implementation details please see the Supplementary Material.

\begin{figure*}
    \centering
    \includegraphics[width=0.9\linewidth]{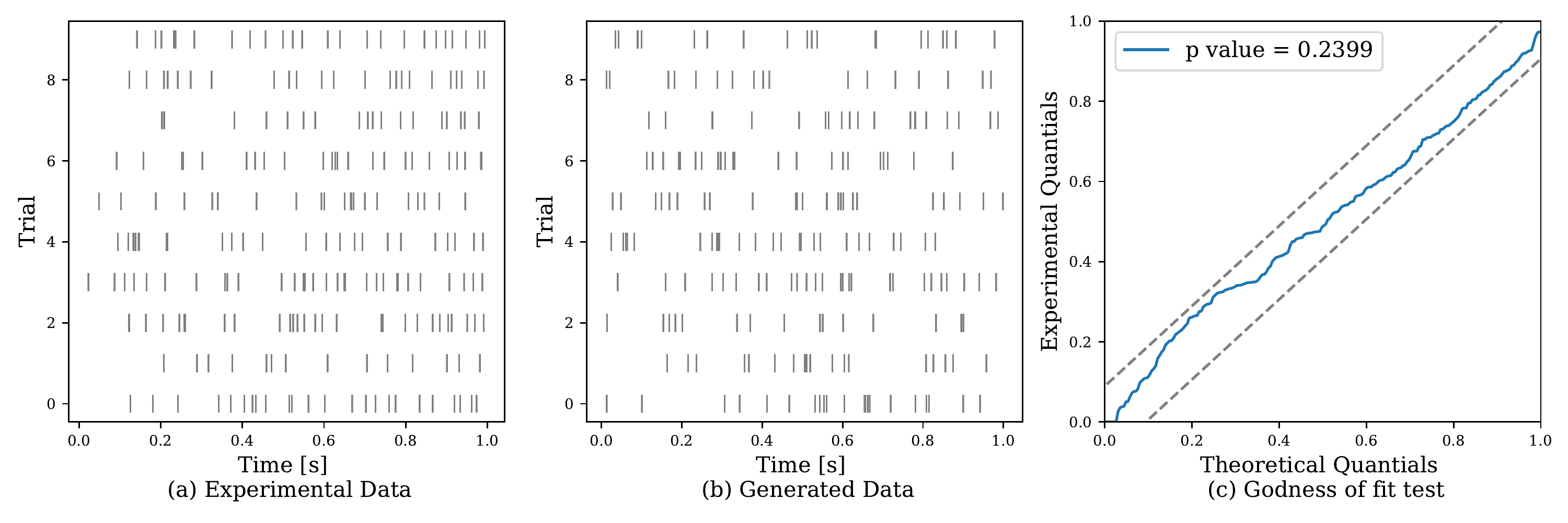}
  \caption{(a) raster plot of a neuron from a monkey cortex. (b) Data generated from the learned model. (c) Results of the Kolmogorov-Smirnov test. The NHGSE generates data that resembles the real data, and passes the goodness of fit test.}\label{fig:neuron_qq}
\end{figure*}

\subsection{Synthetic Data}\label{sec:gen_data}
To assess the performance of the inference algorithms presented in Section~\ref{sec:inference}, we learn the parameters of data generated by the model, and compare the learned parameters to the ground truth. To generate data we start by sampling the memory GP and the background GP, based on Equations~\ref{eq:background_gp} and \ref{eq:memory_gp}. 
We generate events from the model using Poisson thinning \citep{lewis1979simulation}. First we sample the number of candidates $J \sim Poisson\left(\lambda T \right)$, and sample candidate events $\{t_1, \dots t_J\}$ uniformly. Next we chronologically iterate through the candidates and accept them with probability $\frac{\Lambda\left(t_j | \{t_1, \dots t_{j-1} \}\right)}{\lambda}$.

The results for the synthetic data are included in Figure~\ref{fig:gen_dat}. The time window used was one second, and the dataset includes $18$ events. In panel (a), the comparison between the ground truth predictive intensity and the ones inferred by the learning algorithms demonstrates the accuracy of the inference methods. We compare the ground truth to the mean of the Gibbs samples, and the mean of the approximating distribution of the VI. Due to space limitation we do not include the standard deviation of the inferred predictive intensity, and we refer the reader to the Supplementary Material for further results.

Panel (b) compares the ground truth value of the linear intensity and the one inferred by the two learning algorithms. Unlike for the predictive density, the Gibbs sampler is more accurate than the VI. Panel (c) present the inference results for the intensity bound. As expected the approximated distribution by the VI is much more narrow than the distribution of the Gibbs samples.

Panels (d) and (e) show the autocorrelation of the intensity bound, and the ELBO through the Gibbs samples and VI iterations. In this example the convergence of the ELBO is very fast, the autocorrelation of the Gibbs sample vanish only after a few thousands iterations.

We use the test log--likelihood per data point, averaged over ten datasets, to quantify the performance of the two inference algorithms. The Gibbs sampler and the VI achieve very similar results. Thus, in the next section we present the results only from the VI.

\subsection{Real Data}\label{sec:real_data}
\subsubsection{Crime Report Data} \label{sec:crime}
Our model assumes both inhibitory and excitatory self effects, but it should also be able to capture phenomena where only one of the two types of effects exist. To test this, we fit our model to crime report data, where it is assumed that past events have excitatory effect on future events \citep{mohler2011self}. We use the same two datasets described in \citet{zhou2020efficient}, and follow their data processing procedure. Each dataset contains one type of crime and so we use the univariate version of the model. The work of \citet{zhou2020efficient} includes several inference methods and we compare our results to the results of their reported mean--field variational inference approach, as it is the closest to our inference procedure.

Table~\ref{tab:crime} compares the test log--likelihood of our NH-GPS model to the one reported in \citet{zhou2020efficient}. We perform the experiment five times and report the mean and variance of the test log--likelihood. As expected, our model performs similarly to the non--parametric Hawkes process presented by \citet{zhou2020efficient}.

\begin{table}
    \centering
    \caption{Crime Report Data Test Log--Likelihood.}\label{tab:crime}
    \begin{tabular}{ccc}
      \toprule 
      \bfseries Dataset & \bfseries Zhou et al. (2020)  & \bfseries NH--GPS\\
      \midrule 
      Vancouver & $453.11 \pm 8.94$ & $453.8 \pm 12.2$ \\
      NYPD & $-200.7 \pm 3.32 $ & $-202.8 \pm 7.54$ \\
      \bottomrule 
    \end{tabular}
\end{table}

\subsubsection{Neuronal Activity Data}
One of the motivating real world phenomena behind our work is the spiking activity of neurons, where it is known that the process has both self--excitatory and self--inhibitory effects. As an example for our model's ability to capture neuronal activity we use the datasets that were first presented in \citet{gerhard2017stability} (Figure 2.c and 2.b). These data were further analyzed in \citet{apostolopoulou2019mutually} (Figure 5) where the Mutually Regressive Point Process (MR-PP) is introduced. One dataset includes ten recordings from a single neuron in a monkey cortex, with the duration of one second each, and the other includes ten recordings from single neuron in a human cortex for a duration of ten seconds each.

In Figure~\ref{fig:neuron_qq} we assess the ability of the model to capture the data for the recordings from the monkey cortex. The results for the human cortex are included in the Supplementary Material. Panel a and b present the raster plot of the real data and the raster plot generated from the fitted model respectively. Similarly to the real data, the generated data displays both excitation, in the form of clustered events, and inhibition.

To quantify how suitable the model is to the data, we apply the random time change theorem \citep{daley2003introduction} to the inferred intensity and the experimental data. The theorem states that realizations from a general point process can be transformed to realizations from a homogeneous Poisson process with unit rate. Similarly to the work of \citet{apostolopoulou2019mutually}, we further transform the exponential realizations to those from a uniform distribution, following \citet{brown2002time}. We then use the Kolmogorov-Smirnov test to compare the quantiles of the distribution of the transformed realizations to the quantiles of the uniform distribution. The results of this test are displayed in the in Figure~\ref{fig:neuron_qq} c. The comparison relies between the $95\%$ confidence bounds, which are indicated by the dashed lines. The model passes the goodness of fit test (p value $> 0.05$), and the p value is higher than the one achieved by the MR-PP. We compare the reported p--value achieved by the MR--PP model for the two datasets in Table~\ref{tab:neurons}. For both datasets our model achieves higher p--value than the MR--PP.

\begin{table}
    \centering
    \caption{P--Value of the KS Test on Neuronal Activity Data.}\label{tab:neurons}
    \begin{tabular}{ccc}
      \toprule 
      \bfseries Dataset & \bfseries MR--PP & \bfseries NH--GPS\\
      \midrule 
      Monkey Cortex & 0.103 & 0.23 \\
      Human Cortex & 0.096 & 0.175 \\
      \bottomrule 
    \end{tabular}
\end{table}

\subsubsection{Retweets Data}
\begin{figure}
    \centering
    \includegraphics[width=0.4\linewidth]{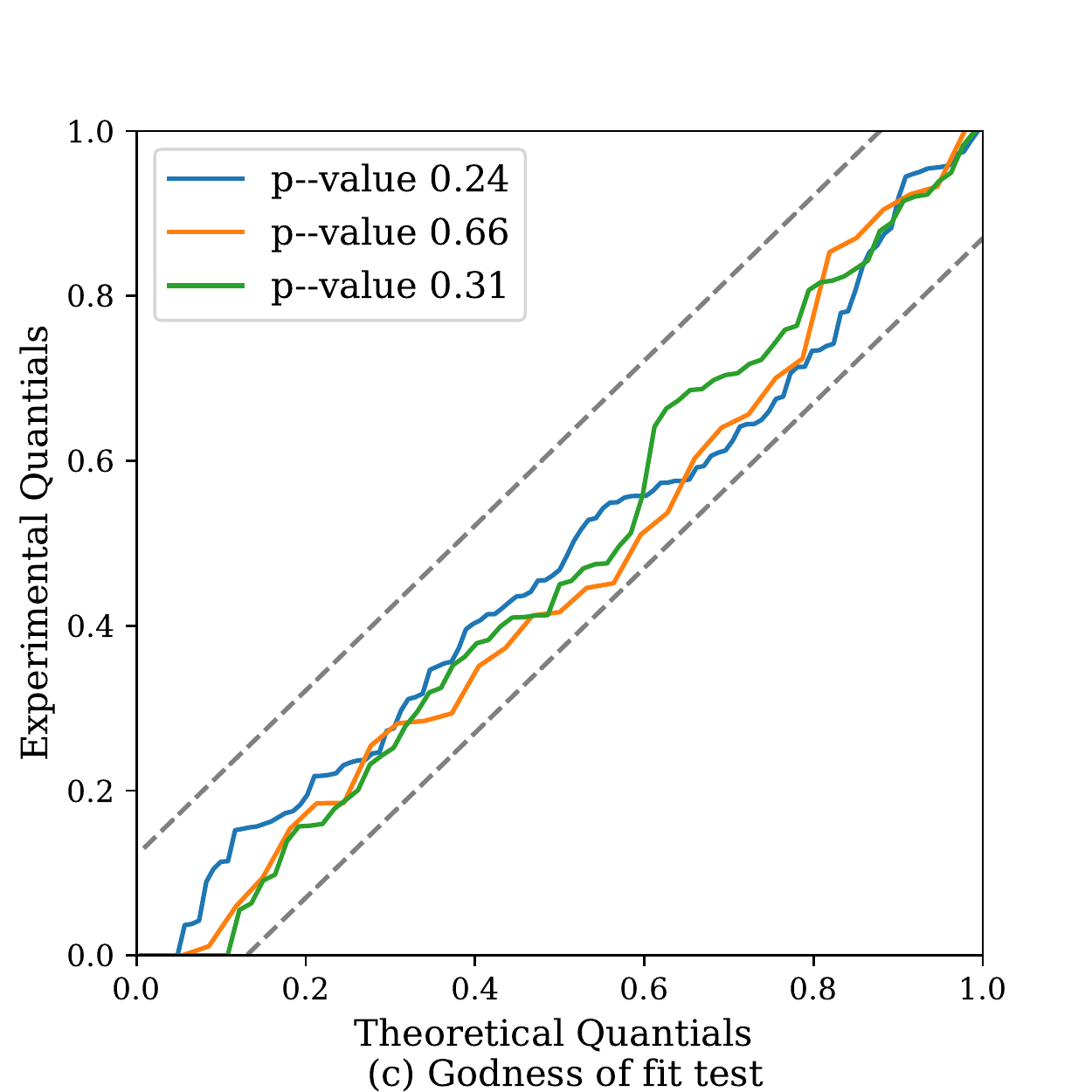}
  \caption{Comparison of the theoretical quantiles and experimental quantiles for each type of event in a retweets cascade dataset.}\label{fig:twitter_qq}
\end{figure}
To demonstrate the necessity of the multivariate variation of the model we use a subset of a retweets dataset. We used the processed data that was published in \citet{zuo2020transformer}. Each time series in the dataset contains the cascade of retweets of a specific tweet and the event time is relative to the time of the initial tweet. Each tweet is tagged according to the amount of followers the user has, and we use the same grouping in to three groups as in \citet{zuo2020transformer}, which results in three event types.

It can be shown that the change time theorem can be applied separately to every dimension of the multivariate intensity function \citep{daley2003introduction}, which in our case corresponds to each of the event types. As our model is best fit to use cases were the data are scarce, we fit a model to each retweets cascade separately. Figure~\ref{fig:twitter_qq} presents the QQ-plot for each type of event for one retweet cascade dataset. All types pass the goodness--of--fit test.

As a baseline we fit the univariate version of the model to the same data, without differentiating between different event types. Table~\ref{tab:twitter} presents the comparison between the fit of the univariate and multivariate versions of the model, averaged over five retweets cascades. It includes the p--values as calculated by the Kolmogorov-Smirnov test, and the test log--likelihood per data point. To calculate the test log--likelihood we trained the model on an hour of retweets and tested it on the following half hour. Both the univariate and the multivariate versions of the model pass the goodness--of--fit test, but the multivariate model achieves a higher p--value, and a higher test log--likelihood.

\begin{table}
    \centering
    \caption{Retweets Data P--Value and Test Log--Likelihood.}\label{tab:twitter}
    \begin{tabular}{ccc}
      \toprule 
      \bfseries Model & \bfseries P--Value & \bfseries Test Log--Likelihood\\
      \midrule 
      Univariate & $0.29 \pm 0.15$ & $-1.99 \pm 0.36$ \\
      Multivariate & $0.58 \pm 0.25$ & $-0.8 \pm 0.6 $ \\
      \bottomrule 
    \end{tabular}
\end{table}

\section{Conclusion}\label{sec:con}
In this work we presented the nonlinear Hawkes model with Gaussian process self--effects (NH-GPS). We motivated the development of the new model with the need for a flexible model that can capture both exciting and inhibiting interactions between events, while maintaining the ability to learn also when data are scarce.

Due to the structure of the model, we dispense with the branching structure that is commonly used for Bayesian inference in Hawkes processes. We propose an efficient mean--field variational inference algorithm which relies on a data augmenting scheme. We show that the results of the variational inference are comparable with those of a Gibbs sampler.

We demonstrate the performance of our model in three different real world applications. Due to the flexibility of our model, it achieves good results on data where events have only excitatory effects and on data where events have both excitatory and inhibitory effects.

Last, we present the multivariate variation of the model. Here, we use it to fit data where events are of different types. This direction can be extended further. The multivariate version of the model can be used to describe events in a network, and even learn the connections in the network. Another future research direction is spatio--temporal processes. The model can be directly extended to capture events in time and space, as the memory kernel is reduced to a kernel function of a GP which can be applied also to multi--dimensional data.

\paragraph{Acknowledgments} The authors would like to thank Dr. Christian Donner for topical discussions which contributed to the research presented in this paper. The research of NMS and MO have been partially funded by the Deutsche Forschungsgemeinschaft (DFG)- Project-ID 318763901 - SFB1294. The research of CO was funded by the BIFOLD-Berlin Institute for the Foundations of Learning and Data (ref. 01IS18025A and ref 01IS18037A).

\bibliographystyle{unsrtnat}
\bibliography{main}  

\newpage
\begin{center}
\textbf{\large Supplementary Materials}
\end{center}

\setcounter{section}{0}
\section{Gibbs Sampler}
We use a blocked Gibbs sampler, which groups two or more variables and sample them at once. Thus, we need to identify the conditional posterior distribution of all the relevant groups.

\subsection{Conditional Distribution of the Upper Intensity Bound}
The conditional distribution of the upper intensity bound is
\begin{align*}
    p\left(\lambda | \{\hat{t}_m, \hat{w}_m\}, \{w_n\}, \phi, \{t_n \} \right) \propto e^{-\lambda T} \lambda^{N + M} p\left(\lambda\right)
\end{align*}
which we identify as a Gamma distribution
\begin{align} \label{eq:lambda_cond}
    p&\left(\lambda | \{\hat{t}_m, \hat{w}_m\}, \{w_n\}, \phi, \{t_n \} \right) \propto Gamma\left(\alpha, \beta \right) \\ \nonumber
    \alpha &= \alpha_0 + N + M \\ \nonumber
    \beta &= \beta_0 + T .
\end{align}

\subsection{Conditional Distribution of the Linear Intensity Function}
The conditional distribution of the linear intensity function in the observed and augmenting events is
\begin{align*}
    &p\left(\phi_{N+M} |\{\hat{t}_m, \hat{w}_m\}, \{w_n\}, \phi, \{t_n \}, \lambda \right) \\ \nonumber 
    &\propto \exp \left(\sum_{n=1}^N f\left(w_n, \phi_n\right) + \sum_{m=1}^M f\left(\hat{w}_m, - \phi_m\right) \right) p\left(\phi_{N+M} \right),
\end{align*}
where we use the shortened notation $\phi_n$ instead of $\phi\left(t_n \right)$ and $\phi_{N+M}$ instead of $\{\{\phi\left(t_n\right)\}_{n=1}^N, \{\phi\left(\hat{t}_m\right)\}_{m=1}^M\}\}$. Given Equation~\ref{eq:f} in the main text, the likelihood term in the posterior is a GP with mean
\begin{align*}
    \mu = \left(\frac{1}{2w_1}, ..., \frac{1}{2w_N}, -\frac{1}{2\hat{w}_1}, -\frac{1}{2\hat{w}_M}\right)^\top
\end{align*}
and diagonal covariance matrix
\begin{align*}
    \Sigma^{-1} = Diag\left(w_1, ..., w_N, \hat{w}_1, \hat{w}_M \right).
\end{align*}
Given the GP prior over $\phi$ the conditional posterior is also a GP
\begin{align} \label{eq:phi_cond}
    &\phi_{N+M} \sim GP\left(\mu_{M+N}, \Sigma_{M+N}\right) \\ \nonumber
    &\mu_{M+N} = \Sigma_{N + M}\Sigma^{-1}\mu \\ \nonumber
    &\Sigma_{N+M}^{-1} = \Sigma^{-1} + \Tilde{K}^{-1}.
\end{align}

\subsection{Conditional Distribution of the PG variables}
The conditional posterior distribution of the augmenting PG variables is
\begin{align*}
    &p\left(\{w_n\}, \{\hat{w}_m\} \right) \\ 
    &\propto \prod_{n=1}^N\exp\left( - \frac{\phi_n^2}{2}w_n\right) \\  
    & \times PG\left(w_n;1,0 \right) \prod_{m=1}^M\exp\left( - \frac{\phi_m^2}{2}\hat{w}_m\right) PG\left(w_m;1,0 \right).
\end{align*}
Using the definition of the tilted PG distribution \citep{polson2013bayesian} the posterior distribution for these parameters is
\begin{align*}
    &w_n \propto PG\left(1, \phi_n \right) \\ 
    &\hat{w}_m \propto PG\left(1, \phi_m \right).
\end{align*}

\subsection{Conditional Distribution of the augmenting events} \label{sec:post_aug_events}
The conditional posterior of the augmenting events is proportional to
\begin{align*}
    p\left(\{\hat{t}_m\}_{m=1}^M | \{t_n\}_{n=1}^N, \{\hat{w}_m\}_{m=1}^M, \{w_n\}_{n=1}^N, \phi, \lambda\right) \\
    \propto \prod_{m=1}^M \lambda e^{f\left(\hat{w}_m, -\phi_m \right)}PG\left( \hat{w}_m ; 1,0\right).
\end{align*}
We recognize the conditional posterior with the unnormalized density of a marked inhomogeneous Poisson process with intensity
\begin{align}
    \Pi_0\left(\hat{w}, \hat{t} \right) = \lambda e^{f\left(\hat{w}_m, -\phi_m \right)}PG\left( \hat{w}_m ; 1,0\right).
\end{align}
The underlying density of the inhomogeneous Poisson process in the realizations space $\mathcal{T}$ is given by
\begin{align*}
    \int_{\mathcal{W}}\Pi_0\left(\hat{w}, \hat{t} \right)d\hat{w} = \lambda \sigma\left(-\phi_m \right).
\end{align*}
To sample the augmenting realizations we use the thinning algorithm \citep{ogata1981lewis}. First, we sample the expected number of events
\begin{align*}
    J \sim Poisson\left(\lambda T\right).
\end{align*}
Next, $J$ candidates are sampled uniformly over the realizations space $\mathcal{T}$. To evaluate the intensity at the candidate points we need to evaluate the linear intensity $\phi$ in these points, given its values in the real events and the previously sampled augmenting events. This can be done using results from GP regression \citep{rasmussen2006gaussian}
\begin{align*}
    &\phi_{J|N+M} \sim GP \left(\mu, \Tilde{K} \right) \\
    &\mu = \Tilde{K}_{J, N+M}\Tilde{K}_{N+M}^{-1}\phi_{N+M} \\
    &\Tilde{K} = \Tilde{K}_J - \Tilde{K}_{J, N+M} \Tilde{K}_{N+M}^{-1} \Tilde{K}_{J, N+M}^\top.
\end{align*}
Once we have $\phi_J$ we perform thinning -  for each candidate $\hat{t}_j$ we generate a random number $r_j$ between $0$ and $1$. If $r_j < \sigma\left(-\phi_j\right)$ we accept the candidate $\hat{t}_j$ and otherwise we discard it.

\begin{algorithm}\label{algo:gibbs}
    \SetAlgoLined
    \KwIn{Observed events $\{t_n\}_{n=1}^N$, hyper parameters $\{ \sigma_g, \alpha, a_s, \sigma_s, \alpha_0, \beta_0 \}$ }
    \KwOut{R samples of $\{\hat{t}_m\}_{m=1}^{M_r}$, $\lambda$, $\phi_{N+M_r}$ }
    Initialize - $M, \phi_{N+M}, \{\hat{t}_m\}_{m=1}^{M_r}$ randomly.
    \For{$r\gets0$ \KwTo $R$}{
        Sample $w \sim PG\left(1, \phi_{N+M} \right)$ \\
        Sample $\lambda \sim \text{Gamma}\left(\alpha, \beta \right)$ as in Equation~\ref{eq:lambda_cond} \\
        Sample $\phi_{N+M} \sim GP \left(\mu_{N+M}, \Sigma_{N+M} \right)$ as in Equation~\ref{eq:phi_cond} \\
        Sample $\{\hat{t}\}_{m=1}^{M_r}$ as in Section~\ref{sec:post_aug_events}
    }
\caption{NH-GPS Gibbs Sampler}
\end{algorithm}

\subsection{Hyper parameters Learning}

The augmented model is not conditionally conjugated with respect to the kernel hyperparameters. This is usually solved by using MCMC within Gibbs sampler approach \citep{gilks1992adaptive, martino2015fast}. This method applies rejection sampling, such as Metropolis--Hastings (MH) \citep{hastings1970monte}, Hamiltonian Monte Carlo (HMC) \citep{duane1987hybrid}, to sample the hyperparameters, and relies heavily on some design choice. A wrong choice of the proposal distribution (for MH) or the mass matrix (for HMC) may result in a very slow convergence, or prevent the sampler from converging at all.

We choose the less traditional approach of taking a gradient step within the Gibbs sampler. This is implemented in the following way -- after sampling all the model parameters from the conditional posterior distributions described above, we derive the negative model log--posterior with respect to the hyperparameters and take a step in the direction of the negative gradient.

This approach can be developed further in the spirit of Stochastic Gradient Descent (SGD). Meaning, rather than updating the hyper parameters after each iteration of the Gibbs sampler, we perform several steps of sampling, take the gradient of the averaged posterior and update the hyper parameters following the averaged gradient.

We include bellow the derivatives with respect for the hyper parameters of the model $\theta = {\sigma_g, \alpha, a_g, \sigma_s}$. 

To learn the hyperparameters  we derive the posterior of the model which appears in Equation~\ref{eq:aug_post} in the main text. First, we notice that all of the hyper parameters appear in the prior over the linear intensity 
\begin{align*}
\log p\left(\phi\right) \propto -\frac{1}{2}\log \det\left(\tilde{K} \right) - \frac{1}{2} \phi^\top \tilde{K}^{-1} \phi
\end{align*}
and all of the hyperparameters appear in the prior kernel.

We next derive an entry in the kernel with respect to the different hyper parameters.
\begin{align*}
    \frac{\partial \tilde{K_{l,k}}}{\partial a_s} &= \exp \left(- \frac{\parallel t_l - t_k \parallel^2}{\sigma_s^2} \right) \\
    \frac{\partial \tilde{K_{l,k}}}{\partial \sigma_s} &= a_s\exp \left(- \frac{\parallel t_l- t_k \parallel^2}{\sigma_s^2} \right) \frac{\parallel t_l - t_k \parallel^2}{\sigma_s^3} \\
    \frac{\partial \tilde{K_{l,k}}}{\partial \alpha} &= \sum_{t_i < t_l}\sum_{t_j < t_k} K_{t_i - t_l,t_j - t_k}^g \\
    &\times \exp\left(-\alpha\left(t_i - t_l + t_j - t_k \right) \right) \left(t_l - t_i + t_k - t_j \right) \\
    \frac{\partial \tilde{K_{l,k}}}{\partial \sigma_g} &= \sum_{t_i < t_l}\sum_{t_j < t_k} K_{t_i - t_l,t_j - t_k}^g \\
    &\times \exp\left(-\alpha\left(t_i - t_l + t_j - t_k \right) \right) \frac{\parallel \left(t_l - t_i\right) - \left(t_k - t_j\right) \parallel^2}{\sigma_g^3}.
\end{align*}
We can plug these results to the chain rule and we get
\begin{align*}
    \nabla \log p\left(\phi\right) = - \frac{1}{2} \text{trace}\left(\tilde{K}^{-1} \nabla \tilde{K} \right) + \frac{1}{2} \phi^{\top} \tilde{K}^{-1} \nabla \tilde{K} \tilde{K}^{-1} \phi .
\end{align*}

\section{Variational Inference}
\begin{algorithm}
    \SetAlgoLined
    \KwIn{Observed Events $\{t_n\}_{n=1}^N$, hyper parameters $\{ \sigma_g, \alpha, a_s, \sigma_s, \alpha_0, \beta_0 \}$ }
    \KwOut{Mean and variance of $\phi$}
    Initialize.
    \While{$\mathcal{L}$ not converged}{
        Estimate the mean and covariance of the GP over the inducing points as in Equation~\ref{eq:vi_gp_ind} \\
        Estimate the mean and variance of the GP over the observations as in Equation~\ref{eq:vi_gp_dat}
        Estimate the mean intensity bound as in Equation~\ref{eq:vi_lambda} \\
        Estimate the mean and variance of the augmenting PG variables as in Equation~\ref{eq:vi_pg} \\
        Estimate the intensity function over the augmenting events as in Equation~\ref{eq:vi_lambda2}
    }
\caption{NH-GPS Variational Inference}
\end{algorithm}
\subsection{Hyper Parameters Learning}
In order to learn the hyperparameters of the model we perform one step of gradient descent with respect to the ELBO at each iteration. The derivatives of the ELBO are give in \citet{donner2018efficient} Appendix F.

\section{Implementation Details}
All the experiments were implemented in Python. To parallelize the computation over the available computing resources we used the JAX package \citep{jax2018github}. In the Gibbs sampler - the sampling of the PG variables was done using the PyP\'{o}lyaGamma package \citep{pypolyagamma}.

\paragraph{Gradient step within the Gibbs sampler}
We update the hyper parameters of the model using SGD. As an optimizer we use the ADAM algorithm \citep{kingma2014adam}. We found that updating the hyperparameters every second iteration yields the best results in terms of running time and convergence time. The ADAM optimizer was also used for the gradient step in the Variational Inference algorithm. We used the recommended default variables of the ADAM optimizer.

\subsection{Hyper parameters values}
The parameters values that were used to generate the synthetic dataset presented in Figure~\ref{fig:gen_dat} are presented in Table~\ref{tab:data_gen}. The hyper parameters for the Variational inference, for all the real world datasets, are presented in Table~\ref{tab:hypers_vi}. The noise level refers to the noise added the diagonal of the relevant covariance matrices.
\begin{table}
    \centering
    \caption{Model Parameters Values for Synthetic Data.}\label{tab:data_gen}
    \begin{tabular}{rl}
      \toprule 
      \bfseries Parameter & \bfseries Value\\
      \midrule 
      Time limit & 1\\
      $\lambda$ & 320\\
      $\sigma_g$ & 0.07\\
      $a_g$ & 1\\
      $\alpha$ & 20\\
      $\sigma_s$ & 0.5 \\
      $a_s$ & 1 \\
      \bottomrule 
    \end{tabular}
\end{table}

\section{Further Results}

\subsection{Synthetic Data}
\begin{figure*}
    \centering
    \includegraphics[width=0.9\linewidth]{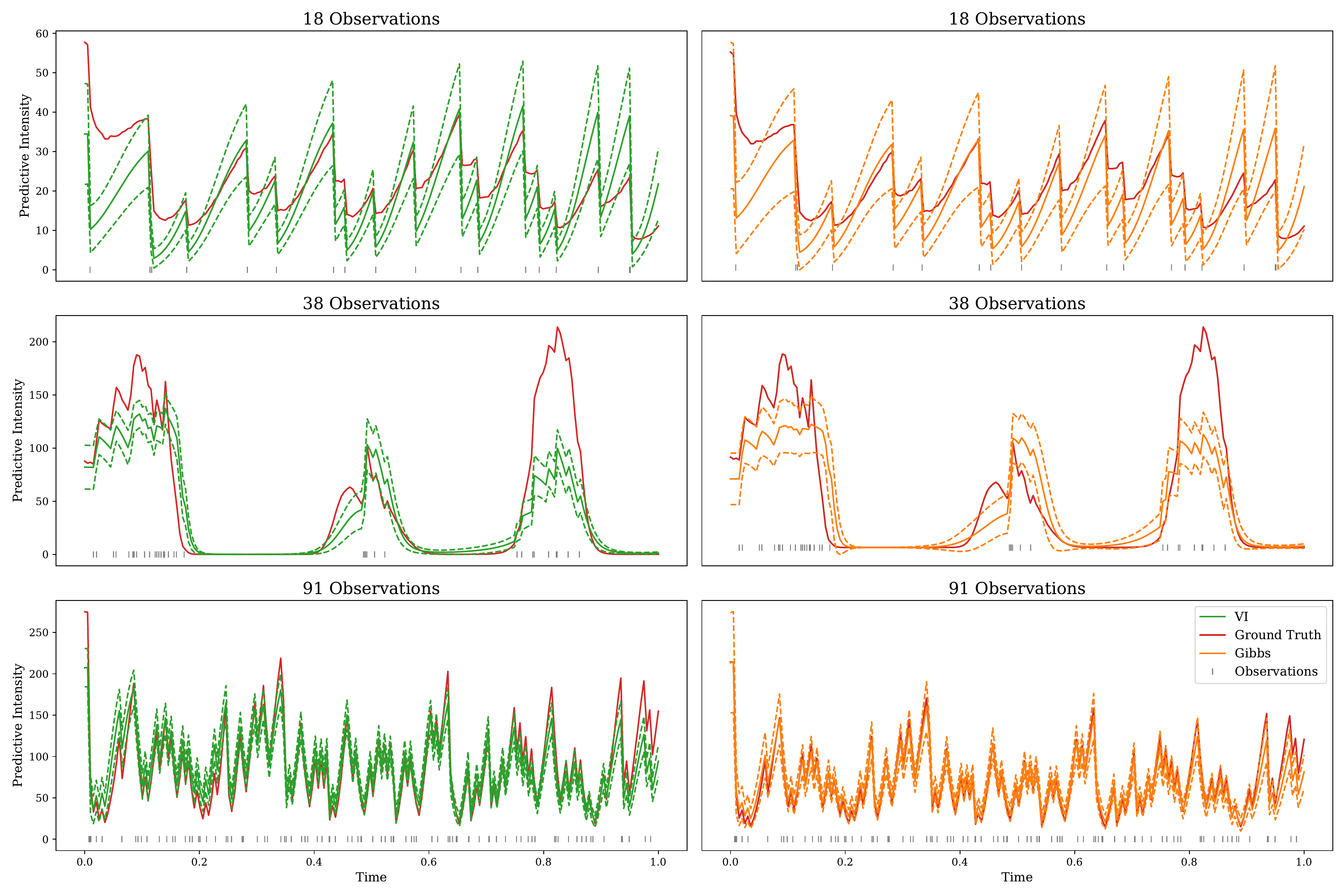}
  \caption{Three examples for the ground truth intensity function and the one inferred by the Gibbs and VI algorithms. The dashes lines are the standard deviation of the inference results.}\label{fig:ints}
\end{figure*}

Figure~\ref{fig:ints} includes three examples of a generated ground truth intensity function and the inference results of the Gibbs sampler and variational inference. The upper panel includes the same data as Figure~\ref{fig:gen_dat} in the main text, but includes also the standard deviation of the inference results.

\subsection{Neuronal Activity Data}
\begin{figure*}
    \centering
    \includegraphics[width=0.9\linewidth]{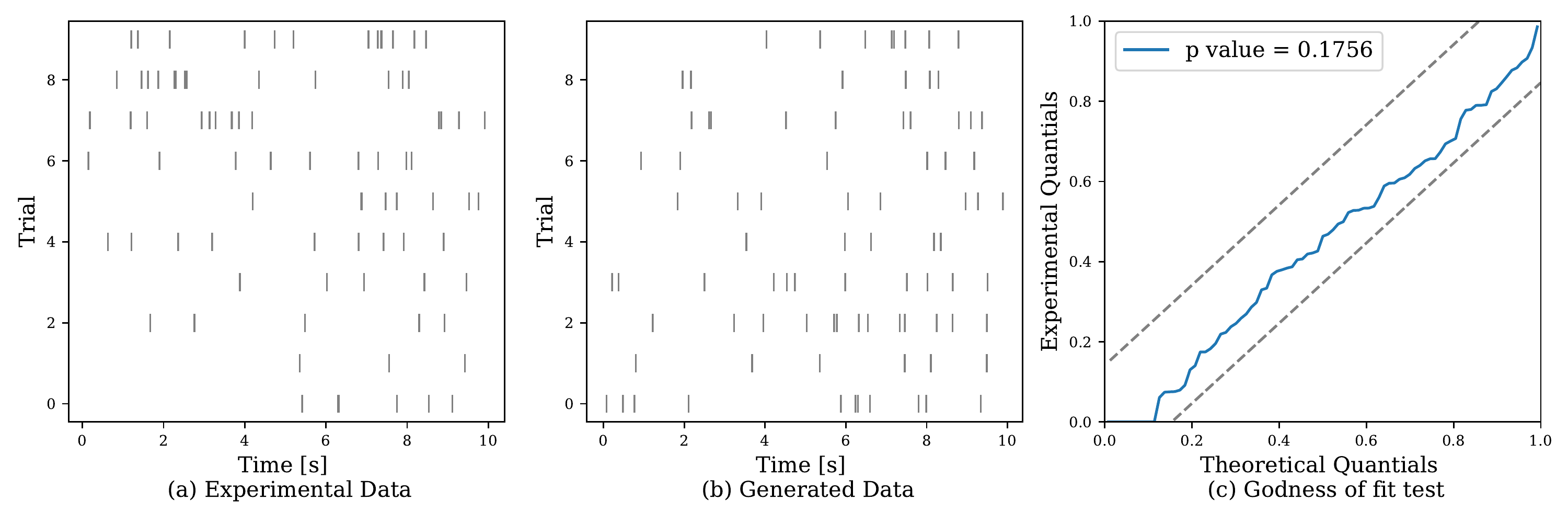}
  \caption{(a) raster plot of a neuron from a monkey cortex. (b) Data generated from the learned model. (c) Results of the Kolmogorov-Smirnov test. The NHGSE generates data that resembles the real data, and passes the goodness of fit test.}\label{fig:neuron_qq_2}
\end{figure*}

Figure~\ref{fig:neuron_qq_2} presents the results for the dataset recorded from a single neuron in a human cortex. Similarly to the results in the main text, the data generated by the model resembles the recordings. Furthermore, the fitted model passes the goodness--of--fit test and the comparison between the theoretical and experimental quantiles are within the $95\%$ confidence bounds.

\begin{table*}
    \centering
    \caption{Hyper Parameters Values for the VI Algorithm.}\label{tab:hypers_vi}
    \begin{tabular}{cccccc}
      \toprule 
      \bfseries Parameter & \bfseries \makecell{Synthetic \\ Data} & \bfseries \makecell{Vancouver \\ Crime} & \bfseries \makecell{NYPD \\ Data} & \bfseries  \makecell{Neuronal \\ Data} & \bfseries \makecell{Retweets \\ Data}\\
      \midrule 
      $\alpha_0$ & $2.5$ & $1.5$ & $1.3$ & $2$ & $1.5$\\
      $\beta_0$ & $0.007$ & $0.1$ & $0.1$ & $0.03$ & $0.1$\\
      \makecell{inducing \\ points} & $100$ & $200$ & $300$ & $100$ & $200$\\
      \makecell{integration \\ points} & $1000$ & $5000$ & $3000$ & $1000$ & $3600$\\
      Noise & $10^{-4}$ & $10^{-4}$ & $10^{-4}$ & $10^{-4}$ & $10^{-4}$\\
      \bottomrule 
    \end{tabular}
\end{table*}

\end{document}